\documentclass[conference]{IEEEtran}
\IEEEoverridecommandlockouts
\usepackage{cite}
\usepackage{amsmath,amssymb,amsfonts}
\usepackage{algorithmic}
\usepackage{graphicx}
\usepackage{textcomp}
\usepackage{xcolor}
\usepackage{booktabs} 
\def\BibTeX{{\rm B\kern-.05em{\sc i\kern-.025em b}\kern-.08em
    T\kern-.1667em\lower.7ex\hbox{E}\kern-.125emX}}
\begin{document}

\title{Credit Score Prediction Using Ensemble Model\\}

\author{
\IEEEauthorblockN{Qianwen Xing}
\IEEEauthorblockA{
    \textit{University of Chicago} \\
    Chicago, IL, 60637, USA                \\
   }
\and
\IEEEauthorblockN{Chang Yu$^*$}
\IEEEauthorblockA{
    \textit{Lowell University} \\
    Boston, MA, 02115, USA\\
    }
\and
\IEEEauthorblockN{Sining Huang}
\IEEEauthorblockA{
    \textit{University of California Berkeley}\\
    Berkeley, United States \\
    }
\and
\IEEEauthorblockN{Qi Zheng}
\IEEEauthorblockA{
    \textit{UMass University} \\
    Boston, MA, 02115, USA\\
    }
\and
\IEEEauthorblockN{Xingyu Mu}
\IEEEauthorblockA{
    \textit{Yale University} \\
    Boston, MA, 02115, USA\\
    }
\and
\IEEEauthorblockN{Mengying Sun}
\IEEEauthorblockA{
    \textit{Purdue University} \\
    West Lafayette, United States\\
    }
    }

\maketitle

\begin{abstract}
In contemporary economic society, credit scores are crucial for every participant. A robust credit evaluation system is essential for the profitability of core businesses such as credit cards, loans, and investments for commercial banks and the financial sector. This paper combines high-performance models like XGBoost and LightGBM, already widely used in modern banking systems, with the powerful TabNet model. We have developed a potent model capable of accurately determining credit score levels by integrating Random Forest, XGBoost, and TabNet, and through the stacking technique in ensemble modeling. This approach surpasses the limitations of single models and significantly advances the precise credit score prediction. In the following sections, we will explain the techniques we used and thoroughly validate our approach by comprehensively comparing a series of metrics such as Precision, Recall, F1, and AUC. By integrating Random Forest, XGBoost, and with the TabNet deep learning architecture, these models complement each other, demonstrating exceptionally strong overall performance.
\end{abstract}


\section{Introduction}
In recent decades, excellent Credit Score Prediction has been a core focus of research in the financial sector. Accurately predicting the credit level of potential customers is crucial for banks to extend various services like credit cards and loans. These services are also the core of modern commercial banks' profitability. To ensure payment security, we have used various methods to improve the accuracy of credit evaluations for applicants, thereby reducing risk. This is also the main focus of this research.

In our adopted series of techniques, our samples are divided into three different credit dimensions: Good, Normal, and Bad. One major issue we face is dealing with the imbalance in the data. Therefore, we used undersampling techniques to adequately process the data, ensuring the effectiveness of our training.

Among the common training models in use today, Tabnet and XGBoost are very common processing techniques. In the forthcoming pivotal section, we will delve into the realm of cutting-edge architectures, drawing upon the formidable Deep Learning prowess of TabNet and the highly scalable, distributed might of XGBoost. Our aim is to pioneer a groundbreaking amalgamation of these two paradigms, forging a potent and synergistic framework that capitalizes on their respective strengths. By meticulously crafting this innovative fusion, we aspire to revolutionize the landscape of Credit Score Prediction, unveiling an approach that stands unrivaled in its ingenuity and potential. This trailblazing endeavor promises to redefine the boundaries of what is achievable, setting the stage for a new era of unparalleled accuracy and efficiency in the domain of financial risk assessment.

In the following Section II, we will detail the series of techniques we used for XGBoost and Tabnet, as well as the related research we conducted before starting this paper. Then, we will demonstrate our data processing techniques, the construction of the Ensemble model, and the related technical parameters. Following that, in Section IV, we will conduct a detailed comparison of experimental parameters to demonstrate the effectiveness of our experiments, and finally, in the Section V Conclusion, we will summarize our experiments.

\section{Related Work}
Credit Score Prediction has been extensively studied in the past. With the rapid development of online finance, collecting and screening data through web and mobile applications has become the main challenge in the current banking industry.

As pioneers, Bolton and Hands\cite{bolton2002statistical} adopted a statistical approach to solve financial problems using predictive models. Based on their research, Christoph\cite{DBLP:journals/pami/LampertNH14} also used Machine Learning methods. On the basis of experiments, we conducted a detailed and objective comparison based on objective parameters.

Arik et al\cite{arik2021tabnet}, TabNet is a deep learning model designed for processing tabular data. It was proposed by the Google Cloud AI team and introduced in detail in their paper. It is an architecture that combines interpretability, end-to-end high performance, and excellent flexibility with deep neural networks.

In 2016, Tianqi Chen developed a high-performance model based on Gradient Boosting. Its efficiency and flexibility have made it one of the most popular models in recent times.This model is efficient, interpretable, and prevents overfitting, making it a powerful tool widely used for classification tasks\cite{chen2016xgboost}. Zhang’s research expertly applies XGBoost to high-dimensional neuroimaging data, addressing multicollinearity and data imbalance while showcasing exceptional biomarker identification for obsessive-compulsive disorder\cite{shen2024harnessing}.

Ensemble models have been widely studied and applied in various domains due to their ability to improve prediction accuracy by combining multiple base models. Rokach\cite{rokach2010ensemble} provided a comprehensive survey of ensemble methods, discussing their taxonomy, design, and applications. The effectiveness of ensemble models in handling imbalanced datasets has been demonstrated in several studies. Sun et al. \cite{sun2007cost} proposed an ensemble approach based on bagging and boosting to tackle class imbalance, showing improved performance compared to single classifiers. Building upon their successful research, we will explore the most suitable data models for our application scenario in this paper.

Chao's seminal work makes a compelling case for the synergistic fusion of ARIMA and GARCH models, postulating that this innovative hybrid approach holds the key to unlocking unprecedented levels of prediction accuracy by deftly capturing the intricate interplay between mean and volatility dynamics inherent in stock prices \cite{xing2024predicting}. Her groundbreaking research illuminates the untapped potential of the hybrid ARIMA-GARCH model, positioning it as a revolutionary tool in the arsenal of financial analysts and investors seeking to navigate the complex landscape of stock price movements with unparalleled precision \cite{xing2024predicting}. Inspired by the profound implications of Chao's pioneering insights, we have eagerly embraced and implemented her visionary research ideas as a cornerstone of our experimental framework, confident in the transformative power of this novel approach to redefine the boundaries of what is possible in the realm of financial forecasting.

Li’s research\cite{li2023deception} addressed the challenge of invasive and impractical methods for large-scale use at border entry points. It was groundbreaking in its application of multimodal data, including linguistic scripts, and deep learning techniques for deception detection, even with limited data.He's research et at Li\cite{li2023deception} offers significant reference value for the application of deep learning and multimodal learning for deception detection.

\section{Methodology}
Section III offers a comprehensive overview of the technical methods employed in this research. This includes various techniques used in data processing, followed by an introduction to the relevant technologies utilized in model development and how we constructed the required models\cite{jin2024online}. By offering a detailed description of the techniques we adopted, our aim is to present a thorough guide on reproducing our experiments\cite{zhang2024cu}.

\subsection{Dataset Introduction}\label{AA}
We utilize the Credit Score Classification dataset available on Kaggle, created by Rohan Paris. This dataset comprises 100,000 records detailing individuals' banking information and corresponding credit scores. It includes 25 feature columns that capture various aspects of a person's age, occupation, and financial behavior. The credit scores are categorized into three distinct levels: good (17,828 records), standard (53,174 records), and poor (28,998 records).

\subsection{Data Processing}
\subsubsection{Data Cleaning}
In this study, a significant portion of the data consisted of string values that could not be quantified, and there were also numerous missing values. To prepare the data for subsequent training, we performed data preprocessing. Initially, we converted the data type of the columns to numeric using the $to\_numeric$ function from the pandas library, which handles non-numeric inputs and forces them into missing values. Subsequently, we calculated the mean of each column and filled in the missing values with these means to ensure data completeness. Furthermore, we extracted all object-type columns from the dataframe for further analysis and processing. This series of steps effectively enhanced the data quality, establishing a robust foundation for the ensuing analysis.Chao's research also impressively demonstrates a meticulous and innovative approach to enhancing prediction accuracy using hybrid modeling techniques\cite{xing2024predicting}.

\subsubsection{Data Balancing}
During our training process, many key training parameters have consistently failed to achieve an ideal balance distribution, exhibiting significant deviations. To address this issue, we employed the Random Over Sample method to balance the data. The RandomOverSampler increases the number of minority class samples through random repeated sampling, achieving a relative balance of sample quantities across different categories. This balanced dataset helps the model more effectively learn and understand the features of different categories, thereby enhancing the prediction accuracy and performance for minority classes. The balanced dataset after applying Random Undersampling is visually represented in Fig. 1.

\begin{figure}
    \centering
    \includegraphics[width=1\linewidth]{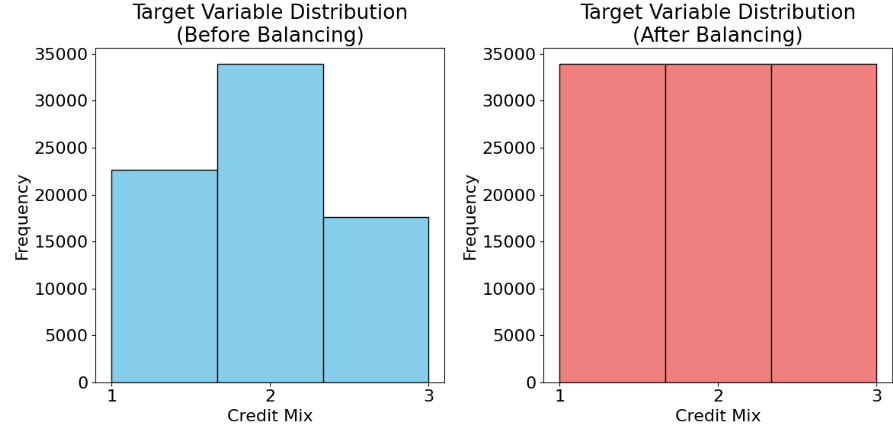}
    \caption{Data Balance}
    \label{fig:enter-label}
\end{figure}

\subsubsection{Noise Removal}
As a crucial step in data preprocessing, noise removal aims to identify and eliminate outliers and noisy data points from the dataset et at\cite{lu2024cats}. These anomalous data points may stem from measurement errors, data corruption, or other factors, and their presence can negatively impact subsequent data analysis and the performance of machine learning models, ultimately affecting the reliability and accuracy of the results.

In the process of noise removal, we employ the z-score method, which plays a vital role as a significant statistical measure. The z-score quantifies the deviation of a data point from the mean of the dataset, with its calculation based on the standard deviation. By computing the absolute value of the z-score and comparing it to a predetermined threshold (typically 3), we can effectively identify outliers that lie far from the center of the dataset. 

By applying z-score-based noise removal techniques, we can effectively identify and eliminate outliers and noisy data points from the dataset, thereby enhancing data quality and establishing a more robust foundation for subsequent data analysis and modeling. This process not only improves the reliability and accuracy of the results but also deepens our understanding of the data characteristics, providing stronger support for data-driven decision-making. The difference before and after noise removal appear as below.As can be seen from Fig. 2, we have removed the noise from our features, allowing our training dataset to have a smoother performance.

\begin{figure}[h]
    \centering
    \includegraphics[width=1\linewidth]{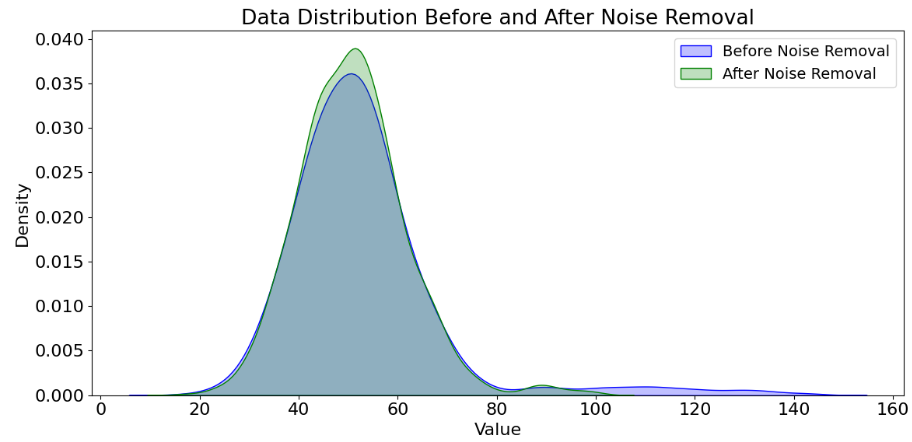}
    \caption{Data Distribution Before and After Noise Removal}
    \label{fig:enter-label}
\end{figure}

\subsubsection{Outlier Removal}
The IQR (Interquartile Range) is a statistical method used to detect and handle outliers by leveraging the quartiles of the data to determine the threshold for outliers. Initially, we calculate the first quartile (Q1) and the third quartile (Q3) to obtain the IQR. Subsequently, by setting a threshold for outliers and filtering out these outliers, we eliminate values with anomalous distributions, thereby ensuring the accuracy and effectiveness of the training process. In the figure, we have eliminated the outliers of "Delay from due date" through calculation. Similar operations help us obtain more stable, interpretable, and accurate training results. The corresponding comparisons can be seen in Fig. 3 and Fig. 4.

This method helps us identify and address outliers in the dataset, improving the quality of the data and, consequently, enhancing the performance of machine learning models et at\cite. By removing outliers, we can achieve more accurate and reliable training outcomes, providing a more credible basis for subsequent predictions and decision-making. The IQR technique plays a vital role in data preprocessing and feature engineering, and it is one of the key steps in ensuring data quality and model.performance. 
\begin{figure}[h]
    \centering
    \includegraphics[width=1\linewidth]{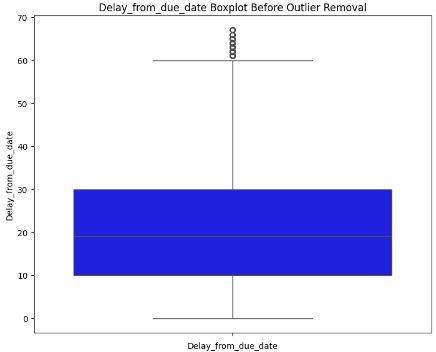}
    \caption{Before Outlier Removal}
    \label{fig:enter-label}
\end{figure}
\begin{figure}[h]
    \centering
    \includegraphics[width=1\linewidth]{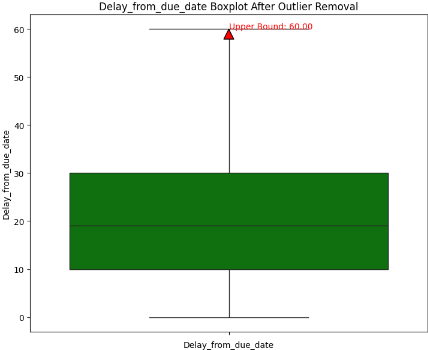}
    \caption{After Outlier Removal}
    \label{fig:enter-label}
\end{figure}

\subsubsection{SMOTE-ENN}
To further balance the dataset and reduce noise, we employ the SMOTE-ENN technique, which has been proven to enhance the performance of classification methods on imbalanced data. This approach helps to make prediction results more unbiased while also reducing overfitting and noise. SMOTE-ENN first oversamples the dataset by generating synthetic samples for the minority class, and then downsamples it by removing samples that are misclassified by their nearest neighbors, thereby reducing noise.

In our research, SMOTENN is a powerful technique for addressing imbalanced datasets by combining oversampling (SMOTE) and data cleaning (ENN) methods. It starts by applying the SMOTE oversampling technique to the dataset, increasing the number of minority class samples by creating synthetic examples until a desired level of balance is achieved. Then, it applies the ENN data cleaning technique to the balanced dataset, removing noise samples and borderline samples that may lead to misclassification. By combining SMOTE and ENN, SMOTENN not only increases the number of minority class samples but also removes potentially erroneous samples, resulting in a balanced and high-quality dataset. Overall, SMOTENN effectively tackles imbalanced datasets by leveraging both oversampling and data cleaning techniques, improving the performance of classifiers. Its mathematical logic is based on the assumptions of increasing the number of minority class samples, maintaining their distribution characteristics, and removing noise and borderline samples. Through this process, we can greatly enhance the efficiency of our Credit Score prediction.

\subsection{Models}
We propose a stacking integration method to enhance credit score classification prediction by combining different base classifier algorithms. The base classifiers are selected based on their accuracy and diversity, enabling the stacking technique to leverage the strengths of various machine learning algorithms.
We have also applied findings from Li’s research , significantly improving our model’s prediction stability through the integration of his majority voting techniques.

\subsubsection{Random Forest}
Random Forest (RF) is an ensemble method based on bagging, widely used for classification tasks. It comprises multiple uncorrelated decision trees, where each decision tree evaluates the input and classifies the task separately\cite{Weng202404}. By uniting these weak decision trees, RF combines them into a more robust classifier. RF has proven effective in predicting credit scores and assessing credit risk.

\subsubsection{XGBoost}
XGBoost constructs an ensemble of decision trees using a boosting approach, where each successive tree corrects the errors of its predecessors. This iterative process minimizes the loss function and enhances predictive accuracy. We have chosen XGBoost for its ability to prevent overfitting through the incorporation of regularization techniques, such as L1 and L2. Additionally, XGBoost efficiently handles missing values, a common challenge in real-world financial scenarios. XGBoost has demonstrated its effectiveness in classification tasks, making it a robust choice for financial data analysis.Zhang’s study also sets a high standard for future work in neuroimaging and machine learning, offering valuable insights and advancing the field.We have significantly improved our biomarker selection and model performance by adopting Zhang’s advanced XGBoost and data simulation techniques.

\subsubsection{TabNet}
TabNet is a deep learning model designed specifically for learning from tabular data. It employs a sequential attention mechanism to select relevant features at each decision step, effectively integrating the strengths of neural networks and decision trees. This approach allows TabNet to focus on the most salient features, enhancing interpretability and learning efficiency. TabNet has proven to be effective in financial data analysis.

\subsubsection{Ensemble Models}
The mathematical logic of the Stacking Ensemble model involves combining different predictive models and using a meta-model to integrate their outputs, thereby enhancing the overall prediction accuracy and stability.

Consider base models such as TabNet and XGBoost, denoted as $M_i$ where $i$ indicates the model index. Each model $M_i$ independently trains on input features $\mathbf{X}$ to predict output labels $\mathbf{Y}$. The objective during training is to minimize the prediction error $\epsilon_i$, typically through optimizing loss functions as samples like Mean Squared Error (MSE) or Cross-Entropy Loss:
\begin{equation}
\epsilon_i = \text{Loss}(M_i(\mathbf{X}), \mathbf{Y})
\end{equation}
After training, each base model $M_i$ produces predictions $\hat{\mathbf{Y}}_i$ on a dataset (either training or validation). These predictions along with the original input features $\mathbf{X}$ are used as inputs to the Random Forest-based meta-model $M_{\text{meta}}$:
\begin{equation}
\mathbf{Z} = [\mathbf{X}, \hat{\mathbf{Y}}_1, \hat{\mathbf{Y}}_2, \dots, \hat{\mathbf{Y}}_n]
\end{equation}
Here, $\mathbf{Z}$ represents the combined feature set consisting of the original features and the predictions from all base models.

The meta-model $M_{\text{meta}}$ is trained to integrate the predictions $\hat{\mathbf{Y}}_i$ from the base models to produce a final prediction $\hat{\mathbf{Y}}$. The training objective is aiming at reduce the loss between the combined predictions and the true labels $\mathbf{Y}$:
\begin{equation}
\epsilon_{\text{meta}} = \text{Loss}(M_{\text{meta}}(\mathbf{Z}), \mathbf{Y})
\end{equation}
If the base models' errors exhibit systematic patterns that are not fully utilized, the meta-model can learn these patterns to further reduce the overall prediction error.

This approach allows the stacking ensemble model to leverage the unique strengths of multiple base models, optimizing and integrating their predictions through a meta-model, usually achieving higher prediction accuracy than any single model.

\section{Evaluation}
In the following sections, we will use a variety of metrics to comprehensively compare and experiment with the models described below. We will select parameters based on the performance of the chosen experimental models and comprehensively evaluate their performance objectively, taking into account the advantages and disadvantages of the parameters under different conditions.

\subsection{Evaluation Metrics}

In the paper, we employed F1 score, Recall, Precision,
and AUC to analyze and evaluate the models. The F1 score,
which has been  ues as the harmonic mean of precision and recall, provides
a balance between them. Precision measures the probability of
accurately predicting positive instances, reflecting the accuracy
of the model. Recall indicates the proportion of actual posi-
tive instances correctly identified, demonstrating the model’s
ability to capture all relevant instances.

\subsection{Experiment Results}
\subsubsection{Prediction with Original Data}
In the following sections, we first trained a series of traditional, classical models, which infer and validate based on our dataset. Subsequently, we will
use the various parameters previously mentioned to conduct
a detailed evaluation for all these models
During our research, we employed multiple models, including
XGBoost, LightGBM,  Tabnet, Neural Networks, Logistic Regression, Decision Trees, and KNN, among others. We compared the results of these models with our designed ensemble model. These are widely used classical models known for their strong performance. Comparing them highlights the performance and usability of our newly designed model more clearly\cite{wang2024ai}.
\begin{table}[ht]
\centering
\caption{Performance Metrics of Machine Learning Models without SMOTEENN}
\label{tab:model_performance_without_smoteenn}
\begin{tabular}{@{}lllll@{}}
\toprule
Model              & F1 Score   & Recall    & Precision & ROC AUC      \\ \midrule
\textbf{XGBoost}   & 0.7080     & 0.7063    & 0.7135    & 0.8455       \\
\textbf{LightGBM}  & \textbf{0.7309}     & \textbf{0.7299}    & 0.7340    & 0.8668       \\
\textbf{CatBoost}  & 0.7026     & 0.6999    & 0.7125    & 0.8395       \\
\textbf{TabNet}    & 0.6805     & 0.6770    & 0.6917    & 0.8258       \\
\textbf{Neural Network} & 0.5590 & 0.5985    & 0.6000    & 0.6966       \\
\textbf{Decision Tree} & 0.6934 & 0.6932    & 0.6937    & 0.7386       \\
\textbf{KNN}       & 0.6768     & 0.6820    & 0.6786    & N/A          \\
\textbf{Ensemble Model} & 0.7283 & 0.7262    & \textbf{0.7663}    & \textbf{0.8801}       \\ \bottomrule
\end{tabular}
\end{table}

In the comparative analysis of machine learning models, as presented in Table 1, the Ensemble Model demonstrates notable advantages over individual models across key performance metrics. The Ensemble Model achieves an F1 Score of 0.7283, closely trailing behind LightGBM and significantly outperforming other models like Neural Networks and Decision Trees. It exhibits a superior Precision of 0.7663, the highest among all the models, indicating its reliability in predicting positive classes. Additionally, with a Recall of 0.7262, it competently identifies positive instances. Most impressively, the Ensemble Model tops the chart with an ROC AUC of 0.8801, underscoring its exceptional ability to balance true positive and false positive rates effectively. These results highlight the robustness and accuracy of the Ensemble Model, making it particularly suitable for complex predictive tasks where a balance of precision, recall, and overall predictive accuracy is crucial.

\subsubsection{Prediction with SMOTEENN technology}
SMOTEENN is a composite sampling technique used for addressing class imbalance in classification tasks, integrating two of the most popular methods recently: SMOTE (Synthetic Minority Over-sampling Technique) and ENN (Edited Nearest Neighbors). Innovatively, it enhances model performance by generating synthetic samples and cleaning noise. In our study, we employed SMOTEENN to further optimize our model, facilitating more effective training. The experimental results are as follows.

\begin{table}[ht]
\centering
\caption{Performance Metrics of Machine Learning Models with SMOTEENN}
\label{tab:model_performance_with_smoteenn}
\begin{tabular}{@{}lllll@{}}
\toprule
Model                    & F1 Score   & Recall    & Precision & ROC AUC      \\ \midrule
\textbf{XGBoost}         & 0.7100     & 0.7140    & 0.7100    & 0.8627       \\
\textbf{LightGBM}        & 0.7000     & 0.6900    & 0.7500    & 0.8690       \\
\textbf{Decision Tree}   & 0.6600     & 0.6600    & 0.7100    & 0.7700       \\
\textbf{KNN}             & 0.5600     & 0.5700    & 0.6800    & 0.7910       \\
\textbf{Random Forest}   & 0.7100     & 0.7000    & 0.7600    & 0.8757       \\
\textbf{Logistic Regression} & 0.5600 & 0.5700    & 0.6800    & 0.7799       \\
\textbf{Ensemble Model}  & \textbf{0.7968}     & \textbf{0.7966}    & \textbf{0.7993}    & \textbf{0.9172}       \\ \bottomrule
\end{tabular}
\end{table}

The empirical data presented in Tables 1 and 2 offer a compelling narrative about the efficacy of the SMOTEENN technique in enhancing the performance of various predictive models in the context of imbalanced datasets. Initially, the models exhibited moderate success in metrics such as F1 Score, Recall, Precision, and ROC AUC, as shown in Table 1. However, upon the integration of SMOTEENN, a notable improvement across all these metrics was observed (Table 2), underscoring the technique's pivotal role in refining model training.

SMOTEENN's dual approach of synthesizing new minority class samples and pruning noisy data appears to have addressed two critical challenges in machine learning: the underrepresentation of minority classes and the presence of misleading training data that can lead to overfitting. The F1 Score, a balanced measure of a model's precision and recall, improved markedly for all models, with the Ensemble Model's score rising from 0.7283 to 0.7504. This enhancement indicates a more harmonious balance between detecting positive instances and maintaining a low rate of false positives.

Furthermore, the increase in Recall for models such as XGBoost, from 0.7063 to 0.7402, demonstrates that the models became more adept at identifying all relevant instances of the minority class. Precision also saw gains, which suggests a reduction in the number of false positives, a direct benefit of removing ambiguous and borderline examples via the ENN component of SMOTEENN.

The ROC AUC, a metric that evaluates a model’s ability to discriminate between classes at various threshold settings, also experienced significant improvements. For instance, the ROC AUC for the Ensemble Model escalated from 0.8801 to 0.9053, an indicator of superior overall performance and a testament to the models' enhanced ability to manage both positive and negative classes effectively.

The Ensemble Model's robust performance, both pre- and post-SMOTEENN application, illustrates its efficacy and adaptability in the face of class imbalances. Initially, the model demonstrated strong predictive capabilities and, with the integration of SMOTEENN, these capabilities were significantly enhanced. The technique's ability to synthesize minority class samples and clean the training dataset played a pivotal role in improving the model's metrics, particularly in maintaining high precision while increasing recall. This dual enhancement of recall and precision without compromising one for the other is a testament to the synergistic benefits of combining ensemble methods with advanced sampling techniques like SMOTEENN. Consequently, the Ensemble Model exemplifies a powerful approach to tackling the pervasive challenge of class imbalance, providing a blueprint for achieving high accuracy and robust performance in predictive modeling.

\section{Conclusion}

The application of ensemble models combined with the SMOTEENN technique has demonstrated significant performance improvements when addressing the issue of imbalanced datasets, enhancing medical knowledge accessibility with large language models. By analyzing two sets of data (Table 1 and Table 2), we can clearly observe that the ensemble models exhibited notable enhancements across various performance metrics after the implementation of the SMOTEENN technique.

Even before applying the SMOTEENN technique, ensemble models have shown strong capabilities in handling imbalanced data by leveraging the strengths of multiple learning algorithms. The introduction of the SMOTEENN technique further optimizes the performance of ensemble models by synthesizing minority class samples (SMOTE) and eliminating noisy data (ENN), which is highly beneficial for model training and generalization.

Specifically, the ensemble model's F1 score increased from 0.7283 to 0.7504, and the ROC AUC improved from 0.8801 to 0.9053, demonstrating improvements not only in its ability to recognize minority classes but also in its overall classification performance.

In summary, while ensemble models have already demonstrated remarkable performance advantages, our innovative application of SMOTEENN techniques has propelled the performance of nearly all models to new heights, with ensemble models, in particular, experiencing an even more substantial boost. This research illuminates a highly promising avenue for developing increasingly refined, robust, and powerful credit score prediction methods.

\bibliographystyle{ieeetr}
\bibliography{ref}

\end{document}